\RequirePackage{xcolor} 
\newcommand{\hilight}[1]{\colorbox{yellow}{#1}}

\newcommand{\eat}[1]{}

\documentclass{article}
\usepackage{soul}
\usepackage{times}
\usepackage{graphicx} 
\usepackage{subcaption} 
\usepackage{tabularx,booktabs}

\usepackage{natbib}

\usepackage{algorithm}
\usepackage{algorithmic}
\usepackage{bm}
\usepackage{array}
\newcolumntype{Z}{>{\centering\let\newline\\\arraybackslash\hspace{0pt}}X}

\usepackage{hyperref}

\usepackage[accepted]{icml2017} 

\icmltitlerunning{Combining LSTM and Latent Topic Modeling for Mortality Prediction}

\begin{document} 

\twocolumn[
\icmltitle{Combining LSTM and Latent Topic Modeling for Mortality Prediction
}

\icmlauthor{Yohan Jo, Lisa Lee, Shruti Palaskar}{\{yohanj,lslee,spalaska\}@cs.cmu.edu}
\icmladdress{}


\vskip 0.3in
]

\begin{abstract} 
There is a great need for technologies that can predict the mortality of patients in intensive care units with both high accuracy and accountability.
We present joint end-to-end neural network architectures that combine long short-term memory (LSTM) and a latent topic model to simultaneously train a classifier for mortality prediction and learn latent topics indicative of mortality from textual clinical notes.
For topic interpretability, the topic modeling layer has been carefully designed as a single-layer network with constraints inspired by LDA.
Experiments on the MIMIC-III dataset show that our models significantly outperform prior models that are based on LDA topics in mortality prediction.
However, we achieve limited success with our method for interpreting topics from the trained models by looking at the neural network weights.
\end{abstract} 

\section{Introduction}
Many intensive care units (ICUs) suffer from a shortage of nurses and doctors to care for patients in critical conditions. As caregivers inevitably have to prioritize patients based on the severity of their conditions, it is essential to leverage patient data---collected from laboratory tests and clinical notes---to help determine the most efficient ICU resource allocation, e.g., by estimating patient mortality.
The problem of mortality prediction involves several challenges.
(1) Mortality prediction requires dealing with time series data, where a progressive analysis of the patients' conditions is preferred to cross-sectional analysis. While prior work has conducted time series analysis of clinical notes \cite{Jo2015,Grnarova:2016ty} and other measurements \cite{Lipton:2015ui} for predicting mortality/diagnoses, there is no standard methodology that yields high accuracy in mortality prediction.
(2) Algorithmic accountability is also critical, as doctors cannot blindly accept the machine's decisions without knowing the rationales behind them.
While neural network techniques have demonstrated promising performance in prediction tasks \cite{Grnarova:2016ty,Lipton:2015ui}, they often lack interpretability and suffer from data sparsity (e.g., text vocabulary), especially in the case of insufficient data.

We present two-layer joint models for mortality prediction that combine the advantages of long short-term memory (LSTM) and latent topic modeling. The LSTM layer captures long-range dependencies in sequential data, trained for mortality prediction, and this information propagates back to the topic modeling layer to learn topics that are predictive of mortality. We tried three different structures for the topic modeling layer. The Encoder only structure has a single-layer network that encodes a bag-of-words to a latent document vector (topic distribution) using a trainable word-topic weight matrix. The Encoder+Decoder structure reconstructs the input vector from the document vector using a trainable topic-word weight matrix that aims to associate each latent dimension with cohesive words. The Encoder+Transcoder+Decoder structure has an intermediate Transcoder layer that converts a document representation to a sparse vector to mimic the sparsity constraints in LDA.

We evaluate our model on the MIMIC-III dataset. The prediction accuracy of our models is compared to that of LDA-based models. The learned topics are evaluated by examining top words associated with each latent dimension. We also analyze the quality of learned topics in terms of their similarity to LDA topics.

\section{Background \& Related Work}

\subsection{Clinical Outcome Prediction}
The three main types of clinical data that have been used for clinical outcome prediction are textual notes~\cite{Ghassemi2014,Jo2015,Grnarova:2016ty}, real-valued measurements (e.g., laboratory/physiologic measurements)~\cite{Lipton:2015ui}, and categorical measurements (e.g., medical codes).

Textual clinical notes contain qualitative information that cannot be found in numeric measurements, such as insights from nurses and doctors, patients' progress, and social context (e.g., relationships with family and friends).  Clinical notes were found to be helpful for long-term prediction, but not as much for short-term prediction~\cite{Jo2015}. Numeric measurements provide useful insight into the patients current health condition and health record. 

Some of the main challenges in analyzing textual clinical notes include unstructured text, incomplete sentences/phrases, irregular use of language, abundant abbreviations, and rich medical jargons and their variations. These characteristics make it difficult to apply NLP tools, such as part-of-speech taggers, dependency parsers, named-entity recognition, etc. Topic modeling is one of the techniques applied to tackle these problems. Ghassemi et al.~(\citeyear{Ghassemi2014}) did a cross-sectional analysis of topics to predict mortality, and later, Jo and Ros\'{e}~(\citeyear{Jo2015}) did a time series analysis using a joint model of HMM and LDA to find patients' latent states and state transition patterns.

Both studies offer good interpretability due to the topics learned from the notes, but the time series analysis with Markovian assumption in the latter study achieved only a minor improvement from the former cross-sectional analysis. Our model can hopefully capture more complex time dependencies in patient trajectories by using an LSTM. Recently, a joint model of LSTM and convolutional neural network (CNN) was used to predict mortality and found phrases that are highly related to mortality~\cite{Grnarova:2016ty}. This model outperforms the previous models, but it is cross-sectional and has limited interpretability. Our work may improve this model by introducing progression analysis and interpretable topics.

For real-valued time series measurements, Chia and Syed~(\citeyear{Chia2014}) predicted mortality using the variability of ECG signals for each patient measured by dynamic time warping between every pair of consecutive heart beats. Chia and Syed~(\citeyear{Chia:2013fw}) also used time series heart rate patterns for mortality prediction, by binning each heart rate (per minute), clustering subsequences of the bins, and choosing clusters that are indicative of mortality and survival respectively. Recently, LSTM without feature engineering was used for diagnosis prediction, where 13 types of time series data (e.g., blood pressure, blood oxygen saturation) were resampled to an hourly rate by taking the mean value and then put into an LSTM for prediction~\cite{Lipton:2015ui}. However, this study found evidence that even simple statistics (e.g., max, min, mean) of the measurements throughout the entire timeline of a patient achieve almost comparable accuracy with a simple MLP.

\subsection{Neural Network for Topic Modeling}
Our goal is to design a neural network architecture, where the upper layer is LSTM for predicting mortality, and the lower layer feeds the LSTM at each time point with a latent topic distribution learned from clinical notes. In this section, we review prior work on modeling topic distributions using neural networks.

One of the simplest and earliest approaches is a restricted Boltzmann machine~\cite{Srivastava:2013wi,Hinton:2009wv}. The activation probability of each hidden node is a sigmoid function on a weighted sum of the frequency of individual words. Hence, a hidden node can be interpreted as a latent topic that has a weight associated with each word. Inspired by these models, a deep sigmoid belief network has been proposed to learn topic distributions in a supervised way given a bag-of-words, and there has been an attempt to interpret the hidden layers~\cite{ZhangLW16}. In the same vein, we try to interpret topics from the hidden layers of our joint models LSTM+E and LSTM+E+D.

In a more recent model, TopicRNN, the topic distribution of text is assumed to be drawn from a normal distribution whose mean and variance are computed from the input BoW using a neural network~\cite{DiengWGP16}. This topic distribution serves as the global semantics of the text, and is fed into an RNN to help predict the following word.

Topics learned by the above models are based on unigrams, but general n-grams may capture better topics for some tasks, as shown in~\cite{Zhai:2013vk,Hardisty:2010ug,Wallach:2006up}. A neural network model has been proposed that learns the association between an arbitrary n-gram and topics~\cite{CaoLLLJ15}. This model, however, takes individual n-grams of interest as separate inputs, which makes training tricky. Recently, hierarchical LSTM has been proposed~\cite{ChungAB16}. Our other joint model, the hierarchical LSTM, exploits this architecture and explores the idea of using a lower LSTM layer for learning topic distributions from a sequence of words without limiting itself to unigrams.



\section{Methods}

\begin{figure}[t]
	\centering
    \includegraphics[width=\linewidth]{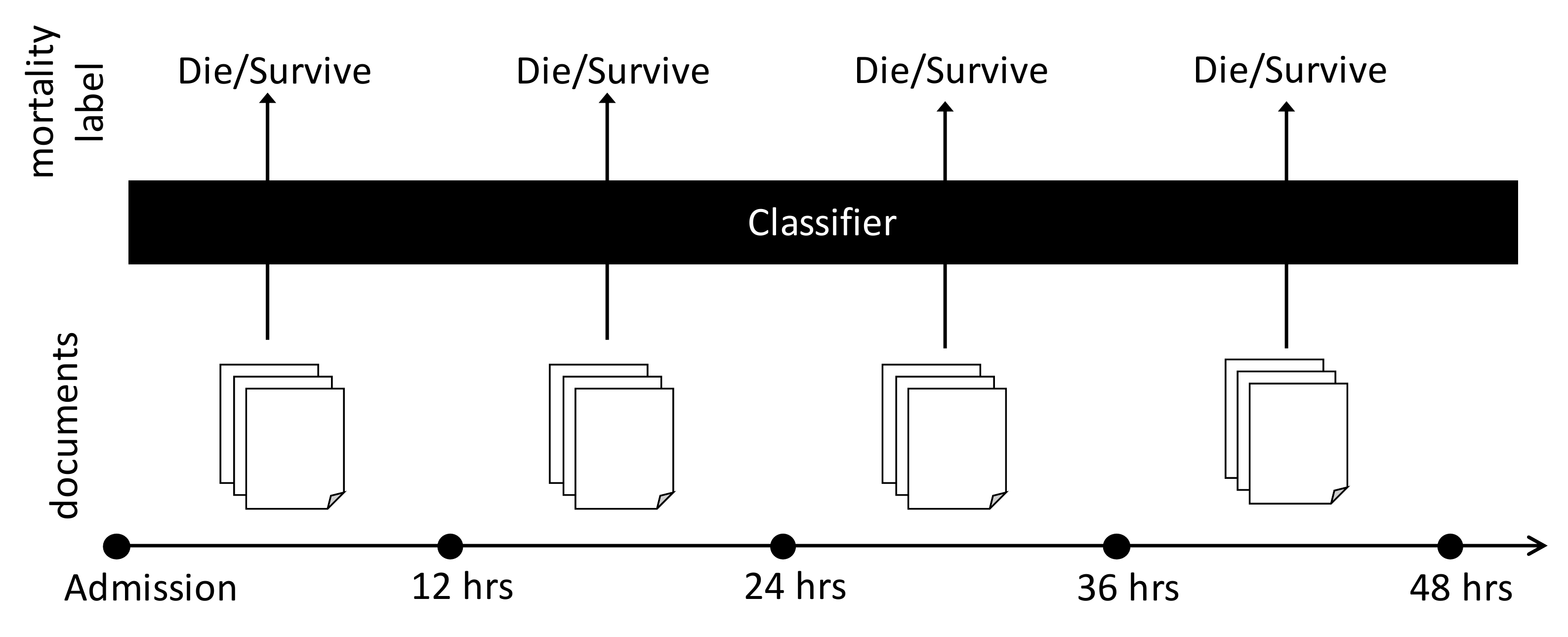}
    \caption{Framework of our mortality prediction task. We define time points as 12 hour-long time segments of a patient's timeline from admission (e.g., time point 1 is the first 12 hours, time point 2 is the next 12 hours, etc.) We aggregate all clinical notes in each time point into one document. At each time point $t$, we use all documents up to $t$ to predict the mortality of the patient at $t$.\label{fig:framework}}
\end{figure}

Our task is to build a classifier for predicting a patent's mortality given his/her clinical notes written so far (Figure \ref{fig:framework}). Throughout the report, we define time points as 12 hour-long time segments of a patient's timeline from admission (e.g., time point 1 is the first 12 hours, time point 2 is the next 12 hours, etc.)~\cite{Ghassemi2014}. For each patient, we aggregate all notes in each time point $t$ into a bag-of-words representation $x_t$, normalized to sum to 1. 

\begin{figure}
\includegraphics[width=0.5\textwidth]{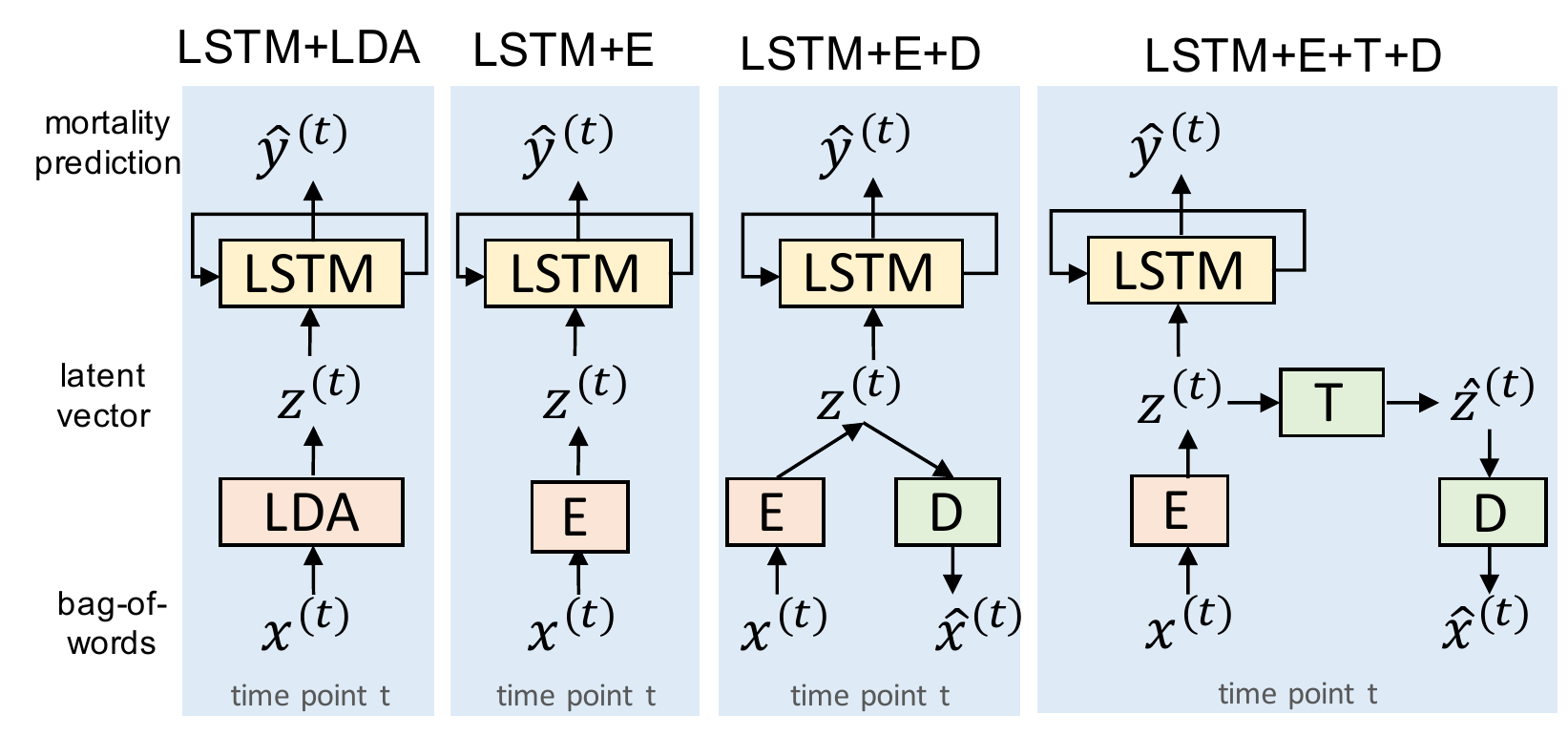}
\caption{Overview of the LSTM models.  In all four models, the input is a bag-of-words representation $x^{(t)}$ of the documents at time point $t$, which is then converted to a latent document embedding $z^{(t)}$ by either a pre-trained LDA model (as in LSTM+LDA) or an Encoder (E) network. The LSTM layer inputs $z_t$ and generates a predicted outcome $\hat{y}^{(t)}$. In LSTM+E+D, we add a Decoder (D) that reconstructs bag-of-words data from $z^{(t)}$. In LSTM+E+T+D, we add an intermediate Transcoder (T) that maps the latent vector $z^{(t)}$ to a sparse topic vector $\hat{z}^{(t)}$, before the Decoder (D) converts $\hat{z}^{(t)}$ to a bag-of-words vector $\hat{x}^{(t)}$. The models are trained to minimize the prediction loss $H(\hat{y}^{(t)}, y)$ and (for LSTM+E(+T)+D) the reconstruction loss $H(\hat{x}^{(t)}, x^{(t)})$ over all time points.
\label{fig:all-models}}
\end{figure}

In this section, we present each of our models (see Fig.~\ref{fig:all-models} for an overview). First, we present and compare two LDA baseline models to analyze the benefit of using LSTM over linear SVM for capturing long-term dependencies (Section~\ref{section:lda-baseline}). Then we present three end-to-end models which jointly learn topic models and mortality prediction (Section~\ref{section:lstm-models}).


\subsection{LDA baselines}
\label{section:lda-baseline}

Here, we present two baseline methods that use LDA, which has been used to infer topic distributions in textual clinical notes. Note that these models are not a joint model of topic modeling and mortality prediction, because they use a separate LDA model to train topics.

{\bf SVM+LDA} is the model proposed by Ghassemi et al.~(\citeyear{Ghassemi2014}). This model builds a linear SVM for each time point that predicts a patient's mortality given the patient's clinical notes written up to that time point. To obtain the input for the classifier for time point $t$, we first compute the topic distribution of each note using LDA and then aggregate all topic distributions up to time point $t$ by simply averaging the topic distribution vectors.

{\bf LSTM+LDA} replaces the linear SVM with LSTM, allowing us to evaluate the ability of LSTM to capture time dependencies for mortality prediction. The LSTM inputs a sequence of topic distribution vectors $\bm{z} = (z^{(1)}, \ldots, z^{(T)})$, generates a predicted outcome $\hat{y}^{(t)}$ at each time point $t$, and is trained to minimize the prediction loss $H(\hat{y}^{(t)}, y)$ over all time points, where  $y \in \{0, 1\}$ is the mortality label of the patient and $H(p, q)$ is the weighted cross-entropy defined as
\begin{equation}
H(p, q) = - C^{FN} q \log p - (1-q) \log (1-p), \label{eq:cross_entropy}
\end{equation}
where $C^{FN}$ is the cost for false negative classification.

Unlike the SVM baseline, LSTM+LDA does not build separate classifiers for different time points, and instead is trained to predict the correct outcome at any time point. By comparing these baselines, we try to establish whether an LSTM can infer relevant semantic information based on topic distributions. (As shown in Fig.~\ref{fig:auc}, LSTM+LDA is more robust than SVM+LDA in predicting the mortality of patients who stay longer in the hospital.)

\subsection{Joint Models}
\label{section:lstm-models}

Next, we present three end-to-end models which jointly learn topic modeling and mortality prediction.

The {\bf LSTM Encoder (LSTM+E)} model replaces the pre-trained LDA topic model in LSTM+LDA with an Encoder (E) network. The Encoder is a single-layer neural network that is expected to find latent topics. Due to the Encoder structure, the Encoder weights $\theta_E$ are reminiscent of LDA word-topic weights in that we can interpret the Encoder as a graphical model that relates words in $x^{(t)}$ to the latent topics in $z^{(t)}$ (see Section~\ref{sec:topic_interpretation} for our analysis).

The Encoder of LSTM+E does not guarantee that cohesive words fall into the same hidden node, which may make hidden nodes difficult to interpret. Hence, {\bf LSTM Encoder-Decoder (LSTM+E+D)} adds a Decoder (D), a single-layer network with no biases that tries to reconstruct the original bag-of-words data by minimizing the reconstruction loss $H(\hat{x}^{(t)}, x^{(t)})$ across all time points. Our rationale behind the Decoder is that the words with the highest weights for each hidden node are likely to be generated together in the same document through the Decoder, thus finding cohesive topics like LDA topics. We can interpret topics from the Decoder weights $\theta_D$ in the same way we do from the Encoder weights.

One potential drawback of LSTM+E+D is that latent document vectors $z$ are tied to both the LSTM network and the Decoder, thereby being optimized neither for mortality nor for topic interpretation. To relax this tie, {\bf LSTM Encoder-Transcoder-Decoder (LSTM+E+T+D)} adds an intermediate Transcoder (T) that maps the latent vector $z^{(t)}$ to a sparse topic vector $\hat{z}^{(t)}$, before the Decoder (D) converts $\hat{z}^{(t)}$ to a bag-of-words vector $\hat{x}^{(t)}$.


Inspired by LDA's sparsity constraints on the document's probability distribution over topics and the topic's probability distribution over words, we impose L1 regularizations on the learned topic vector $\hat{z}^{(t)}$ and the Decoder's weights $\theta_D$, respectively. Thus, our final loss function for LSTM+E+T+D is
\begin{equation}\label{eq:autoencoder-loss}
\sum_{t=1}^T \left[ \underbrace{H(\hat{y}^{(t)}, y)}_{\textrm{\footnotesize prediction}} + \lambda_1  \underbrace{H(\hat{x}^{(t)}, x^{(t)})}_{\textrm{\footnotesize reconstruction}} + \lambda_2 \| \hat{z}^{(t)} \|_1 \right]  + \lambda_3 \| \theta_D \|_1
\end{equation}
where $\lambda_1, \lambda_2, \lambda_3$ are hyperparameters to control the weight of each loss.

Unlike LSTM+LDA, which uses a pre-trained topic model to generate topic vectors for documents at each time point, these joint LSTM models are end-to-end in that they jointly learn the topic models and the contexts between documents. As a result, they learn a better latent document representation $z^{(t)}$ for mortality prediction and outperform LSTM+LDA on all three mortality prediction tasks (see Fig.~\ref{fig:auc}).

\section{Experiments}

\subsection{Evaluation Metrics}

Following Ghassemi et al. (\citeyear{Ghassemi2014,Grnarova:2016ty}), we evaluate our model on three mortality prediction tasks: in-hospital mortality, 30-day post-discharge mortality, and 1-year post-discharge mortality. These three tasks cover both short-term and long-term mortality.
We measure accuracy via the Area Under ROC Curve (AUC) metric~\cite{Rakotomamonjy:2004ts}, which captures how well a trained classifier discriminates between positive instances and negative instances.


\subsection{Data and Preprocessing}
\label{section:data}
We use the MIMIC-III
dataset which contains data about patients admitted to critical care units at a large tertiary care hospital~\cite{Johnson:2016km}. 
For data preparation, we followed Ghassemi et al. (\citeyear{Ghassemi2014})'s work as closely as we could, as their setting has been widely used~\cite{Jo2015,Grnarova:2016ty}. We also excluded patients who are less than 18 years old; these patients (mostly infants) show very different trajectories from adult patients~\cite{Jo2015}. 
We used all textual clinical notes except discharge summaries because they explicitly mention the patient's outcomes. We randomly split patients into training, validation, and test sets with the ratio of 6:2:2. Since the classes are severely skewed toward negative (i.e., survival), the negative instances in the training set are downsampled such that negative instances constitute no more than 70\% of the training set. The validation and test sets are not downsampled.

For preprocessing of the text, we first normalized some text into categories to cluster meaningful text. We replace deidentified information in text with the given category (e.g., ``[** First Name 3 **]'' $\rightarrow$ ``\#\#firstname\#\#''). We also replace times and numbers with ``\#\#time\#\#'' and ``\#'' using regular expressions. Next, we tokenized the text with non-alphanumeric letters. We removed stop words using the Onix stop word list\footnote{\url{http://www.lextek.com/manuals/onix/stopwords1.html}}.

To reduce the vocabulary size, we retained at most 500 words that have the highest tf-idf among all documents for each patient in the training set, and only included these words into the vocabulary. We excluded subjects from the training set if the total length of the clinical notes is less than 100 words. The statistics of the final data after preprocessing are listed in Table \ref{tab:data_stats}.

For our LSTM models, we handle missing time points by using zero vectors for documents $x^{(t)}$. Note that this setting does not affect the prediction and reconstruction loss.

\begin{table}
	\begin{footnotesize}
	\centering
    \begin{tabularx}{\linewidth}{lrXr} \toprule
    \# patients & 36,218 & \# unique words & 83,176 \\ 
    Seq. len (median) & 13 & Seq. len (max) & 1,145 \\ 
    Doc. len (median) & 113 & Doc. len (max) & 2,507 \\ \bottomrule
    \end{tabularx}
    \caption{Data statistics for MIMIC-III after preprocessing. Here, ``Seq. len'' refers to the last time point of a patient's timeline from admission, and ``Doc. len'' refers to the number of words in the concatenation of the notes at a time point $t$. \label{tab:data_stats}}
    \end{footnotesize}
\end{table}

\subsection{Models and Parameters}

For the LDA baseline, the number of topics is set to 50~\cite{Ghassemi2014}. The cost $C$ and the weight $w$ for the positive class (``died'') in libsvm are explored on the grid of $C = \{2^{-5},2^{-3},2^{-1},2^{1},2^{3},2^{5},2^{7},2^{9},2^{11},2^{13},2^{15}\}$ and $w = \{1,3,5,7,9\}$ and tuned on the validation set. The weight for the negative class (``survived'') is fixed to 1.


\begin{table*}[ht]
\begin{tabularx}{\linewidth}{p{5cm}ZZZZ} \toprule
 & LSTM+LDA & LSTM+E & LSTM+E+D & LSTM+E+T+D \\ \midrule
\# nodes in the topic layer & 50 & 50 & 50 & 50 \\
\# nodes in the LSTM hidden layer & 128 & 128 & 128 & 128 \\
LSTM activation & Softmax & Softmax & Softmax & Softmax \\ 
Encoder activation & - & ReLU & ReLU & ReLU \\
Transcoder activation & - & - & - & ReLU \\
Decoder activation & - & - & Softmax & Softmax \\ 
\bottomrule
\end{tabularx}
\caption{Network configurations for each of our models.\label{tab:network_config}}
\end{table*}

The network configuration of our models is summarized in Table \ref{tab:network_config}. The batch size is 10, the number of training steps is 100,000, and the learning rate is 0.001. As in the SVM, we explored different false negative costs (Eq. \ref{eq:cross_entropy}) $C^{FN} = \{2^0, 2^1, 2^2, 2^3\}$ and chose the optimal value based on the validation set.

For the loss function (Eq \ref{eq:autoencoder-loss}), we explored $\lambda_1 = \{10^{-2}, 10^{-1}, 10^0, 10^1\}$ and $\lambda_2, \lambda_3 \in \{0, 1\}$. Ultimately, we found that $\lambda_1 = 10^{-2}$, $\lambda_2 = 0$, $\lambda_3 = 1$ resulted in the highest AUC score for mortality prediction on the validation set, and used these settings for our experiments.

\section{Results}
\label{model-results}

We evaluate our models on both mortality prediction (Section~\ref{sec:mortality-prediction}) and the quality of learned topics (Sections~\ref{sec:topic_interpretation} and~\ref{sec:topic_quality}).

\subsection{Mortality Prediction Accuracy}\label{sec:mortality-prediction}

\begin{figure*}[ht]
\begin{subfigure}[t]{0.33\linewidth}
	\includegraphics[width=\linewidth]{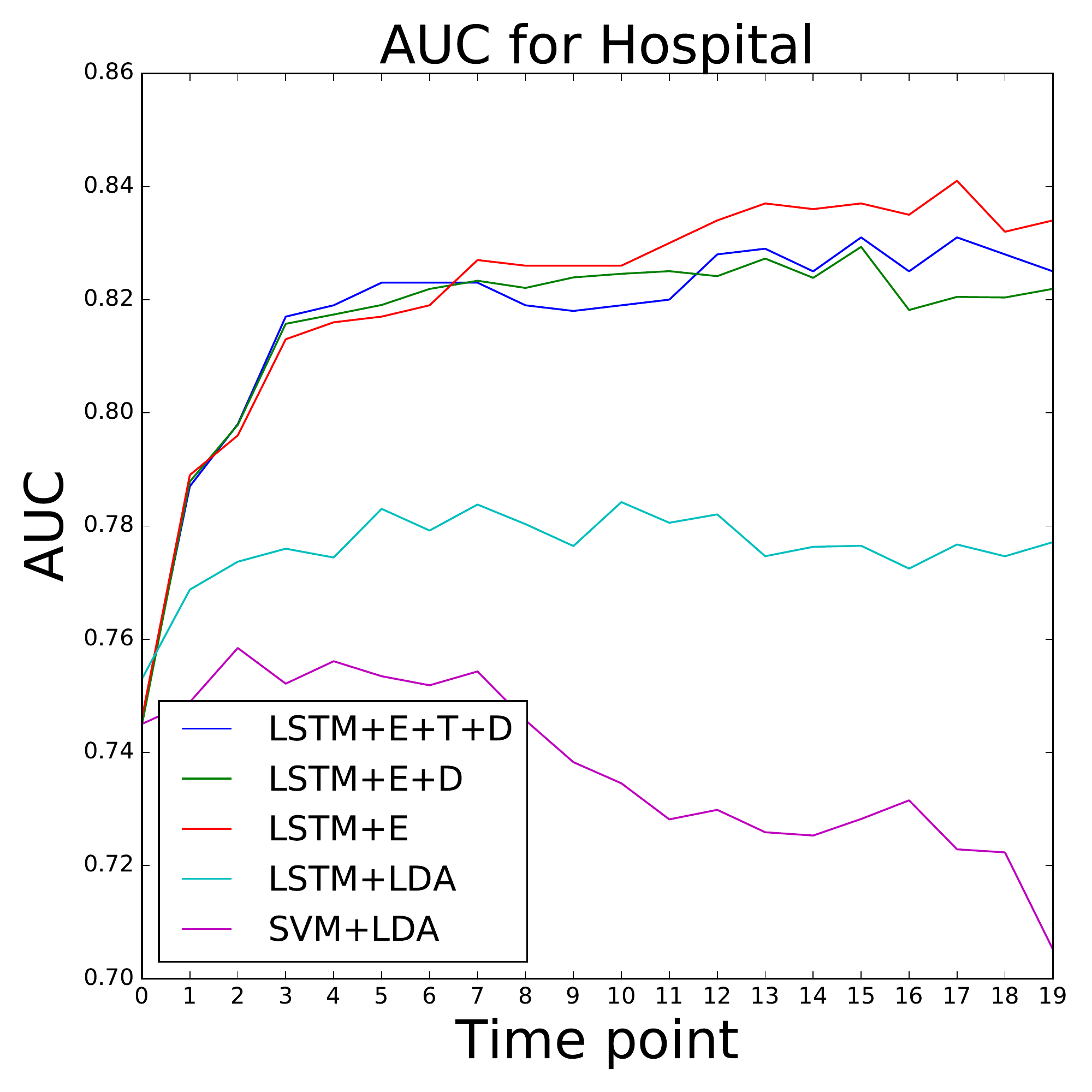}
\end{subfigure}
\begin{subfigure}[t]{0.33\linewidth}
	\includegraphics[width=\linewidth]{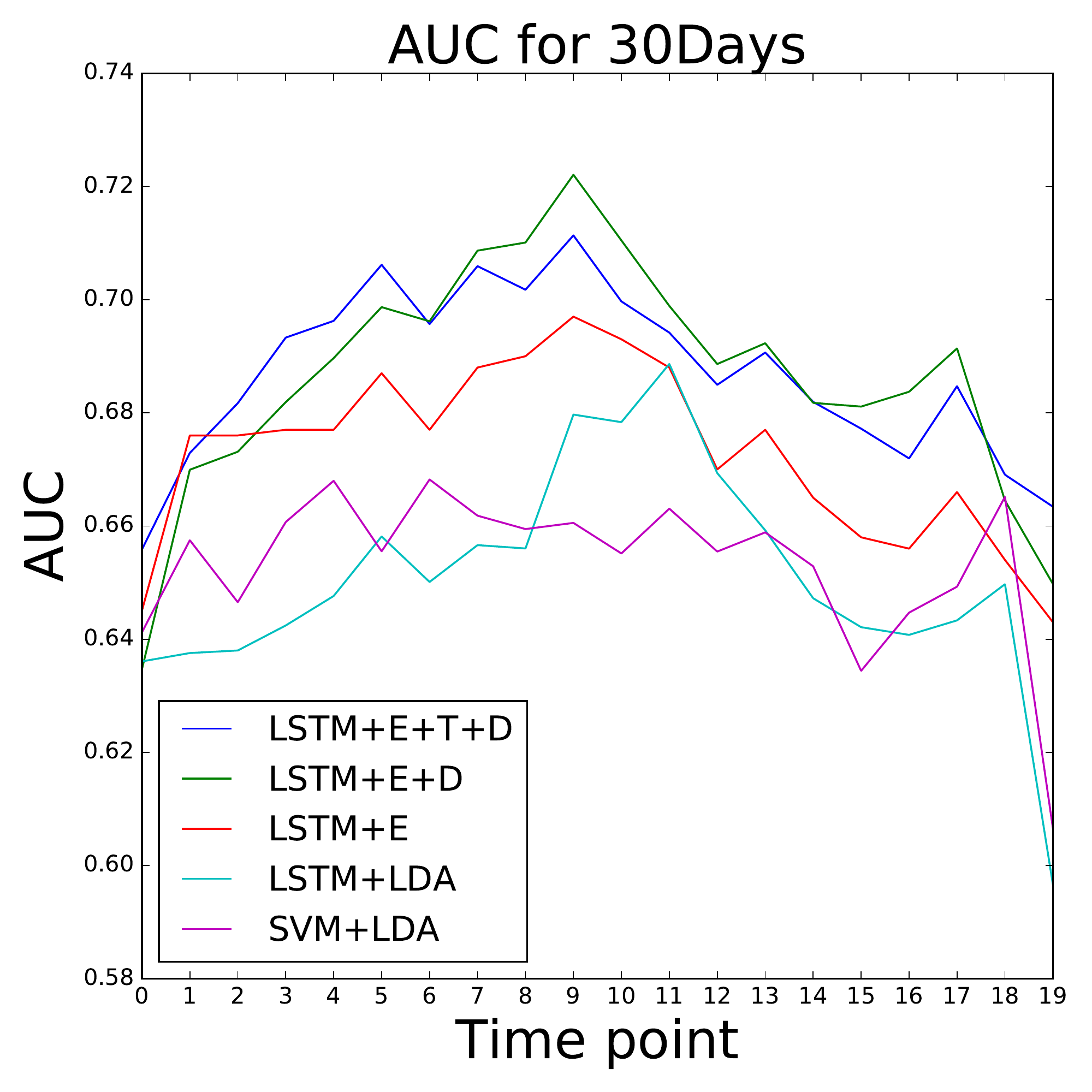}
\end{subfigure}
\begin{subfigure}[t]{0.33\linewidth}
	\includegraphics[width=\linewidth]{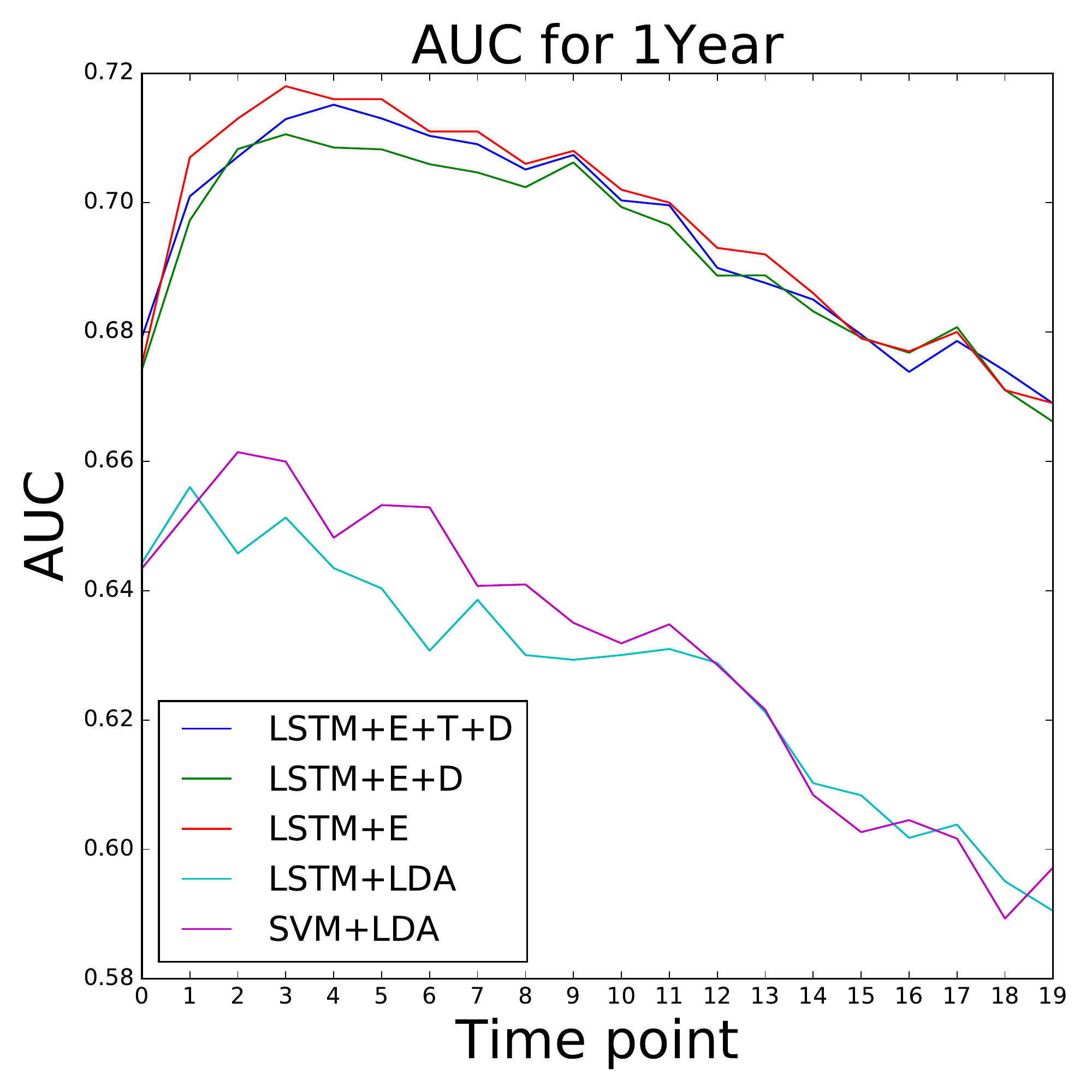}
\end{subfigure}
\caption{Accuracy on predicting in-hospital mortality (left), 30-day post-discharge mortality (center), and 1-year post-discharge mortality (right) at every time point. We use the Area Under ROC Curve (AUC) metric. The LSTM+E(+D) models outperform the LDA baselines on all three mortality prediction tasks.\label{fig:auc}}
\end{figure*}

In Fig.~\ref{fig:auc}, we plot the AUC score of each model for each time point on the three mortality prediction tasks. The joint models LSTM+E(+T)(+D) outperform the LDA baselines on all three mortality prediction tasks. For example, LSTM+E outperforms LSTM+LDA by about +4\% (Hospital), +2\% (30Days), +7\% (1Year) on average. 

The LSTM-based approaches show less accuracy drop over time than the SVM+LDA baseline, notably for the in-hospital prediction. One of the reasons for SVM+LDA's performance drop is having fewer training instances for later time points, as patients who have died or been discharged at a certain time point are excluded from the training set~\cite{Ghassemi2014}. According to the in-hospital prediction, LSTM seems to be able to relieve this problem. We suspect that the way we define our cost function---the average of losses at previous time points---might have compensated for the data loss. Another reason for performance drop is that it is more difficult to predict the destiny of patients who stay in an ICU for long. Long-term time dependencies captured by LSTM might have help make stable prediction.

We introduced the cost $C^{FN}$ for false negative classification to compensate for the fewer number of positive instances, but this cost had inconsistent effects. For example, we found that a cost greater than 1 improved the accuracy for in-hospital and 1-year post-discharge mortality prediction, but decreased the accuracy for the 30-day post-discharge prediction.

\begin{figure*}[t]
\includegraphics[width=\textwidth]{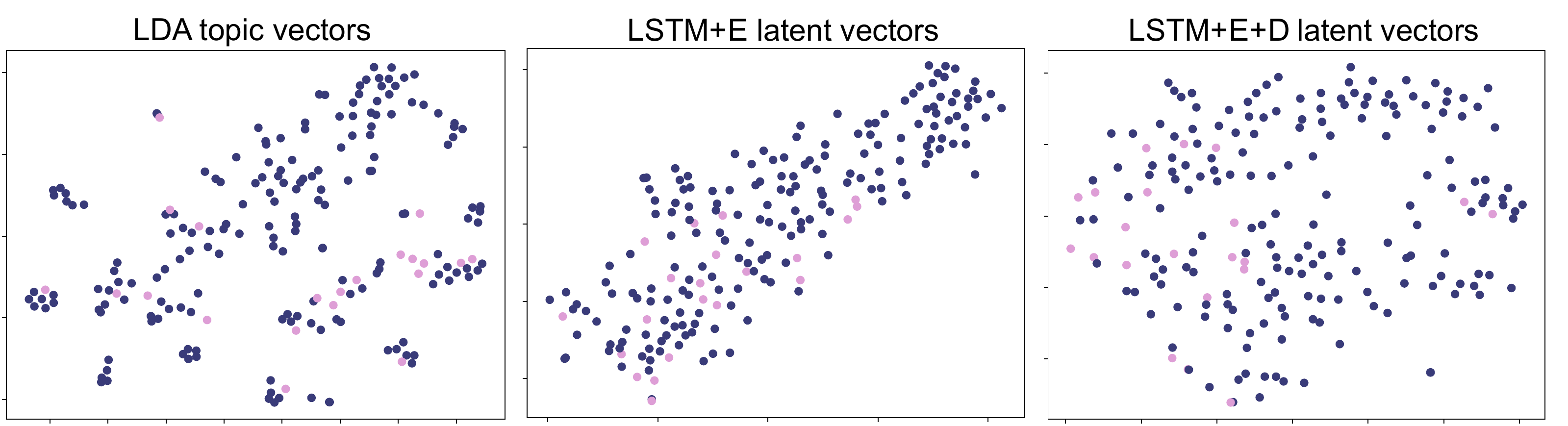}
\caption{We visualize how well the latent representations separate documents according to mortality label. Each dot corresponds to a latent document embedding $z^{(t)}$, colored by in-hospital mortality label where pink indicates positive instances (``died'') and dark blue indicates negative instances (``survived''). The joint models LSTM+E(+D) produce latent representations that better separate documents corresponding to mortality rate: the pink dots are more closely clustered together in the ``LSTM+E(+D) latent vectors'' plots, whereas the pink dots are more spread apart in the ``LDA topic vectors'' plot.
\label{fig:tsne-mortality}}
\end{figure*}

The latent document vectors learned by our joint models have a better ability to separate documents by mortality rate. This is demonstrated in Fig.~\ref{fig:tsne-mortality}, where we visualize the latent document vectors $z^{(t)}$ for each model, and see that the joint models, LSTM+E(+D), have documents with the same mortality label clustered more closely.

\begin{figure*}
\includegraphics[width=\textwidth]{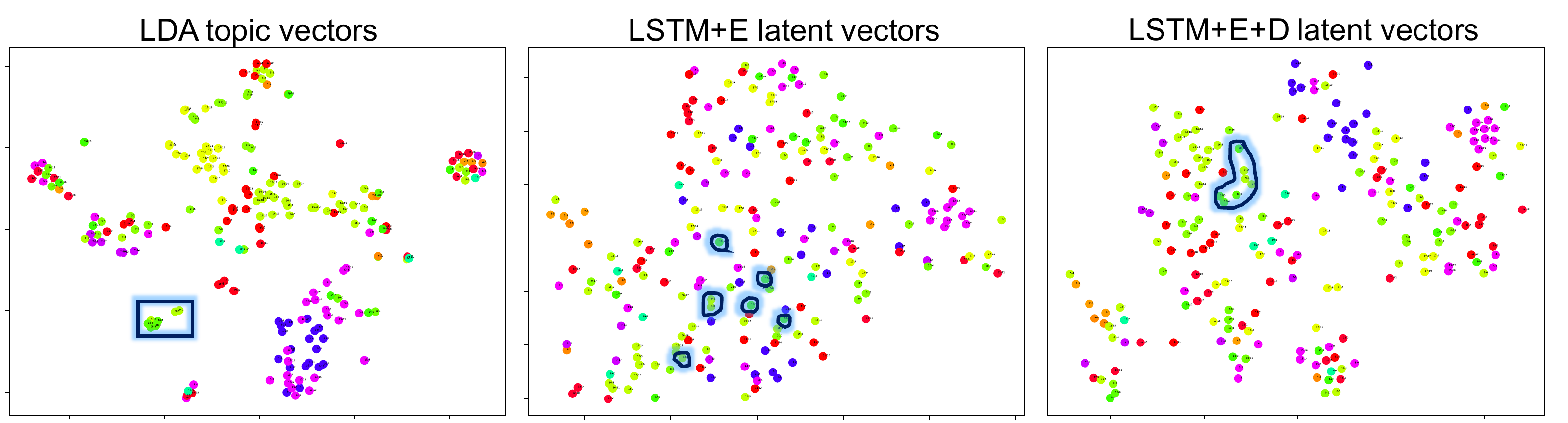}
\caption{
We visualize the topic clusterings produced by the latent vectors of each model. Each point represents to a document, and each color corresponds to a different patient. (We colored the dots by patient id because documents belonging to the same patient are more likely to have the same topics, and so it is slightly easier to visualize topic clusters.) Generally, we observed that the Autoencoder model (LSTM+E+D) produces topic clusterings that are more similar to the LDA topic clusterings than the Encoder model (LSTM+E). For example, consider the cluster of documents inside the box in the ``LDA topic vectors'' plot. We circle the corresponding documents in the LSTM+E(+D) plots. Notice that in the LSTM+E+D plot, the documents are also clustered closely together, but in the LSTM+E plot, these documents are more dispersed. We observed this general trend for many LDA topic clusters, and concluded that the Decoder network indeed helps the model learn better topic distributions. We did not observe any significant difference in the quality of topic clusters learned by LSTM+E+D vs. by LSTM+E+T+D, so we omit the plot for LSTM+E+T+D here.
\label{fig:tsne-topics}}
\end{figure*}

\subsection{Topic Interpretation\label{sec:topic_interpretation}}

\begin{footnotesize}
\begin{table*}[ht]
\begin{subtable}{\linewidth}
  \begin{tabularx}{\linewidth}{cX} 
	T0 & \#, pt, impaired, \#\#date\#\#, activity, sit, status, balance, mobility, stand  \\
	T1 & \#, \#\#date\#\#, \#\#time\#\#, am, dl, mg, meq, weight, arterial, nutrition  \\
	T2 & tube, \#, placement, chest, reason, line, \#\#date\#\#, tip, \#\#time\#\#, left  \\
	T3 & \#, insulin, gtt, blood, dm, hr, diabetes, patient, continue, type  \\
	T4 & chest, \#\#date\#\#, reason, ap, \#, portable, left, \#\#time\#\#, examination, report  \\
  \end{tabularx}
  \caption{Top 10 words with the highest weights in the LDA model\label{tab:topics_lda}}
\end{subtable}%
\vspace*{4mm}

\begin{subtable}{\linewidth}
  \begin{tabularx}{\linewidth}{cX} 
	T0 & barium, dsfx, diagonosis, darts, ndka, swallon, palplabe, cpapwith, ileoileo, laminar \\
	T1 & evenign, tranversing, \#dex, bronchusplan, ecxema, degenerate, flunisolide, sangiunous, sternotony, interstitices \\
	T2 & thorcic, medisternal, proxtmal, merry, reslts, depressant, presacral, godchild, killed, q\#am \\
	T3 & lb\#, wenckebach, quaderant, aticoagulated, nonproducted, emg, en, hypotherm, faa, dalaudid \\
	T4 & diagnosing, gallstones, retroardiac, parieto, lymphoproliferative, isoview, pati, femoroacetabular, periampullary \\
  \end{tabularx}
  \caption{Top 10 words with the highest Encoder weights $\theta_E$ in LSTM+E\label{tab:topics_encoder}}
\end{subtable}%
\vspace*{4mm}

\begin{subtable}{\linewidth}
  \begin{tabularx}{\linewidth}{cX} 
	T0 & tacromulius, kerly, parathroid, programming, coure, placeemrnt, toilteting, reveals, intervan, pheochromocytoma \\
	T1 & filmr, dap, nugauze, extubed, intracystic, plsn, replete, implements, intraparotid, wrote \\
	T2 & sunburn, infrahilar, peppermint, extenisve, \#reisman, intraconal, overwhich, isoview, evevning, ischiorectal \\
	T3 & finesteride, helmut, rhabdomyo, hepatopulmonary, displ, comparison\#, franc, doudoerm, posterioly, diminutive \\
	T4 & relan, interluminal, reinserted, peforation, reticulation, ooob, allohsct, cholangiograms, hemiparesus, contniued \\
  \end{tabularx}
  \caption{Top 10 words with the highest Decoder weights $\theta_D$ in LSTM+E+D\label{tab:topics_decoder}}
\end{subtable}%
\vspace*{4mm}

\begin{subtable}{\linewidth}
  \begin{tabularx}{\linewidth}{cX} 
	T0 & recomendations, left, odorous, tachynea, \#pcs, doxazosin, specifice, hydromyelic, message, spiration \\
	T1 & pharmacology, pancreati, cofirm, frn, eventially, wthdrawal, rslts, wean, famiily, rpb \\
	T2 & relaxation, baseclinical, titarted, sadk, lymphomatoid, bph, mahogany, atelelectasis, frontalis, dlr \\
	T3 & carcino, eea, orlsca, hospitization, captain, suringe, wellness, obstructed, agrestat, dilatation \\
	T4 & after\#, syggestive, nutritions, leni, diruetics, diaphroetic, vase, breathsounds, insominia, perirectal \\
  \end{tabularx}
  \caption{Top 10 words with the highest Decoder weights $\theta_D$ in LSTM+E+T+D\label{tab:topics_transcoder}}
\end{subtable}
\caption{Example topics T0, $\ldots$, T4 and their top 10 words sorted by LDA weights (a) or the trained networks weights (b-d). The topics were chosen randomly from LDA and from the joint models trained on in-hospital mortality. Limited interpretability is offered by these top words, which include many typos (e.g., ``evenign'', ``placeemrnt'') and rare words or medical jargons (e.g., ``tachynea'', ``doxazosin'').\label{tab:topics}}
\end{table*}
\end{footnotesize}

A popular method for qualitatively analyzing latent topics is to examine the top words of each topic. 
We expected each hidden node in the Encoder to represent a cohesive topic that is indicative of mortality. We tried interpreting the hidden nodes by examining the words that have the highest weights for each hidden node. As shown in Table \ref{tab:topics_encoder}, the top words consist largely of typos and infrequent words and thus hardly represent an interpretable notion of topics compared to LDA topics in Table \ref{tab:topics_lda}. We conclude that (I) mortality is related with certain words individually rather than as a group like an LDA topic, and (II) the Encoder weights cannot find cohesive topics partly because words in the same document are not encouraged to be tied to the same hidden node.

We introduced the Decoder of LSTM+E+D to alleviate (II). We tried interpreting topics from the Decoder weights in the same way we did from the Encoder weights. As shown in Table \ref{tab:topics_decoder}, the top words consist of more frequent words than do the top words from the Encoder weights. Yet, there still appear many typos and infrequent words, and the topics are quite difficult to interpret. This does not necessarily mean that topics are not cohesive. Due to the limited time, we could not further investigate the learned topics, which consist of a lot of medical terms and jargons. We leave this for future work.

The topics interpreted from the Decoder weights of LSTM+E+T+D are shown in Table \ref{tab:topics_transcoder}. We expected that the sparsity constraints for $\hat{z}^{(t)}$ and $\theta_D$ imposed on this model would produce more interpretable, cohesive, LDA-like topics, but the learned topics do not show significant difference from those from LSTM+E+D. Moreover, we found that L1 regularization on $\hat{z}^{(t)}$ caused the AUC to fall, while L1 regularization on the Decoder weights $\theta_D$ increased the AUC. However, these results might be due to using a suboptimal setting for topic modeling.  We leave further experiments on finetuning $\lambda_1, \lambda_2, \lambda_3$ in Eq. (\ref{eq:autoencoder-loss}) to balance the tradeoff between better mortality prediction performance and better latent topic modeling as future work.

Interpreting topics from single-layer encoder and decoder turns out to be extremely difficult. The Encoder seems to pick important words predictive of mortality, instead of finding cohesive topics. The structure of the Decoder is similar to probabilistic latent semantic indexing (PLSI), which computes the probability of word $w$ in document $d$ as follows:
$$p(w|d) = \sum_{t} p(w|t) p(t|d),$$
where $p(w|t)$ and $p(t|d)$ correspond to the input vector and the weights of the Decoder, respectively. We imposed sparsity on $p(w|t)$ and $p(t|d)$ in an attempt to make topics more interpretable, but we could not attain satisfactory topics. In order to make sure that the Decoder works as expected, we may need to run PLSI on the data and compare the result topics with those obtained from our models. The learned topics may not be interpretable simply because documents are noisy; each document is a concatenation of multiple clinical notes in the same time point that may be of very different types.

\subsection{Topic Quality\label{sec:topic_quality}}

We evaluate the quality of learned topics by analyzing the t-SNE plots of the latent document vectors in Fig.~\ref{fig:tsne-topics}. We visualized the LDA topic vectors and the latent vectors of LSTM+E(+D) with colors representing patients (Fig.~\ref{fig:tsne-topics}). We focused on the documents in the blue box in the left pane and looked at whether these documents are clustered closely in the LSTM+E (middle pane) and LSTM+E+D (right pane) plots. LSTM+E+D seems to produce a more similar clustering to LDA than LSTM+E does, and thus LSTM+E+D may outperform LSTM+E with respect to the first method. Note that we did not observe any significant difference in the quality of topic clusters learned by LSTM+E+D vs. by LSTM+E+T+D, so we omitted the plot for LSTM+E+T+D.



\eat{
The results are summarized in Table \ref{tab:knn}.

\begin{table}[ht]
  \begin{tabularx}{\linewidth}{cZZZZ} \toprule
  	 & \multicolumn{2}{c}{LDA Topics} & \multicolumn{2}{c}{Patient IDs} \\ \cmidrule[.2pt](lr){2-3} \cmidrule[.2pt](lr){4-5}
   	 & $k=5$ & $k=10$ & $k=5$ & $k=10$ \\ \midrule
   	LSTM+D & & & &  \\
   	LSTM+E+D & & & &  \\
   	LDA & - & - & &  \\
   \bottomrule
  \end{tabularx}
  \caption{Topic quality measured by two gold standard clusters (LDA topics and patient IDs) with $k$ nearest neighbors. \hilight{TODO(shruti)} \label{tab:knn}}
\end{table}
}





\section{Conclusion}

We presented three joint models of LSTM and topic modeling that combine the benefits of both (1) LSTMs, which can capture time dependencies for mortality prediction, and (2) topic modeling, which can interpret topics predictive of mortality. Our models improved the accuracy of mortality prediction significantly from the baseline LDA-based models and were able to learn latent document representations indicative of mortality. We also proposed a method for interpreting topics from our models based on the Encoder and Decoder weights. However, the words with the highest weights for each hidden node did not provide interpretable, cohesive topics as LDA topics do. This may imply that LDA-like topics are suboptimal as feature for mortality prediction, and more indicative information is conveyed rather by certain individual words. We tried to make the Decoder work in a similar manner to PLSI and LDA by imposing constraints, but it turned out to be extremely difficult to obtain interpretable LDA-like topics from the Decoder. This noise might come from our concatenating multiple clinical notes of different types in the same time point.

\textbf{Future work} 

To understand which words are indicative of mortality and survival, we can investigate which words trigger each hidden node in the LSTM. To do this, we calculate the derivative of the value of each hidden node in the LSTM cell in terms of each input word, while fixing the weights of the whole network after they are trained. By training in this way, our model may be able to generate a bag-of-words vector that maximizes each hidden node in the LSTM.

We observed that most of the top words for each hidden node (ranked by the Encoder or Decoder weights) are typos (e.g., ``recomendations'', ``evenign'') or rare words. To make these top words more interpretable, we suggest preprocessing the data in the following ways: (1) Remove the $n$ most frequent and infrequent words. (2) Use a spell-checker to fix typos. Since electronic health records have many medical jargons, we want to be careful and only replace 1-character typos (e.g., correct ``cooeprative'' to ``cooperative'').

In Section~\ref{sec:topic_quality}, we evaluated the quality of learned topics by analyzing the t-SNE plots of the latent document vectors. For future work, we suggest a method to quantitatively evaluate the learned topics by comparing the clusters they form to a gold standard.
More specifically, for each latent vector, we identify $k$ nearest neighbors and compute the overlap with gold standard neighbors of the vector. Since no ground truth quality scores are available, we tried two gold standards: LDA topics and patient identity. In the first method, we identify the $k$ nearest neighbors for each LDA latent vector and compute the overlap with the $k$ nearest neighbors of the same document vector for LSTM+E and LSTM+E+D, respectively. Document representations similar to LDA topics may be considered good in that they represent cohesive, interpretable topics, but setting the standard to LDA topics is not optimal, as our ultimate goal is to find more indicative topics than LDA topics. In the second method, we assume that each patient has consistent topics, and for each latent vector, we compute the percentage of the $k$ nearest neighbor vectors that are from the same patient. The assumption may not hold, however, if a patient's symptoms and condition are diverse and change significantly over time. 

Since we interpret topics from the Decoder weights $\theta_D$ in LSTM+E(+T)+D, we can use arbitrarily deep Encoder and Transcoder networks to increase the complexity of our model, which may allow the model to learn better hidden representations for both topic modeling and mortality prediction. However, keep in mind that a more complex network requires more training data in order for the model to saturate.


\bibliography{main}
\bibliographystyle{icml2017}

\end{document}